\documentclass[smallcondensed]{svjour3}       
\smartqed  
\usepackage{subfigure}
\usepackage{pifont}
\usepackage{geometry}
\usepackage{txfonts}

\usepackage{amsmath,amsfonts, graphicx}
\usepackage[T1]{fontenc}
\usepackage{ulem}
\usepackage{array}
\usepackage{float}
\usepackage{amssymb}
\usepackage{chngpage}

\usepackage[figuresright]{rotating}

\begin{document}

\title{A unifying approach on bias and variance analysis for classification}
\author{Cemre Zor*  \and Terry Windeatt}

\institute{*Corresponding author \\ \\ C. Zor \at
              Centre for Medical Image Computing (CMIC) \\
              University College London \\
              London, WC1E 7JE, UK
              \email{c.zor@ucl.ac.uk}           
           \and
           T. Windeatt \at
               Centre for Vision, Speech and Signal Processing (CVSSP) \\
               University of Surrey \\
               Guildford, GU2 7XH, UK 
               \email{t.windeatt@surrey.ac.uk}
}

\maketitle

\begin{abstract}
Standard bias and variance (B\&V) terminologies were originally defined for the regression setting and
their extensions to classification have led to several different models / definitions in the literature.
In this paper, we aim to provide the link between the commonly used frameworks of Tumer \& Ghosh (T\&G) and James.
By unifying the two approaches, we relate the B\&V defined for the $0/1$ loss to the standard B\&V of the boundary distributions given for the squared error loss. The closed form relationships provide a deeper understanding of classification performance, and their use is demonstrated in two case studies.

\keywords{Bias and Variance \and Classification \and  Performance Analysis \and  Regression}
\end{abstract}

\section{Introduction}

Bias and variance (B\&V) analysis is one of the main approaches that
provide useful insights into the underlying theories
behind classification performance. The analysis is borrowed from the
regression setting and aims to decompose the prediction error of a
given classifier into the terms of B\&V to evaluate their effects
on the performance. Therefore, it can help answer questions such as
``How can we compare the accuracy of two different types of classifiers?'',
``What is it that makes stronger classifiers perform well? Is it
the reduction in the bias they bring about, or in variance, or both?''.
Other than being theoretically  interesting, the answers
to these questions are also meant to provide  better classifier design
strategies which bring about improved prediction performance.

After the initial decomposition of the prediction error into the
\textsl{standard B\&V} terms in the regression setting by  \cite{DBLP:journals/neco/GemanBD92},
different studies have attempted to carry over
this analysis into the classification setting while preserving the meanings of
the terms and the additive property of the decomposition.
These studies had to address two main difficulties:  1) In classification, the predictor (estimator)  responses
(labels/targets) are obtained as hard outputs instead of the soft/continuous responses of regression  2) The associated loss function is $0/1$ in contrast
to the squared error. In order to circumvent the difficulties, different frameworks have proposed significantly different ways to define  \textsl{the classification B\&V}, yet,
almost all frameworks have brought about multiple drawbacks. Some examples for these drawbacks can be given as the
limitations on the use of these frameworks with general loss functions other than $0/1$,
and the existence of impractical and non-standard characteristics like
negative variance and non-zero bias assigned to the Bayes classifier.

Standing out from the rest of the frameworks, \cite{DBLP:journals/ml/James03} and \cite{DBLP:conf/aaai/Domingos00}
have overcome most shortcomings and managed
to provide coherent decompositions applicable for general loss functions while adhering to the characteristics
of the standard B\&V of the regression setting. In particular, James has defined a generalised B\&V in line with the original  meaning, such that, bias quantifies the average distance between the aggregate predictor
output and the aggregate data response, and the variance quantifies
the variability of the predictor. James has also proposed two new notions, systematic/bias
effect (\textit{SE}) and variance effect (\textit{VE}), in order to retain the additive
decomposition of the error for all loss functions. 

From a different point of view,
\cite{DBLP:journals/pr/TumerGOriginal,TumerGhoshCorrelated,tumerGhoshSon} have assumed the existence of soft classifier outputs and directly inherited the B\&V terms from the regression setting rather than proposing new definitions. In other words, they have shown that it is  possible to decompose the classification error into the standard B\&V of the estimation errors belonging to the underlying probability distributions, using squared error loss. 

In the literature, there have been studies focusing on
the relationship between the various definitions of classification
B\&V by formulating them in unified frameworks \cite{DBLP:journals/datamine/Friedman97,DBLP:journals/ml/James03,Kuncheva:2004:CPC:975251}.  However, there has not yet been any attempt to link the  Tumer \& Ghosh (T\&G) analytical model to any other classification B\&V  framework. The aim of this paper is to establish this missing connection by relating  T\&G  to James' model,
 chosen as a representative of the classification  B\&V frameworks.  Specifically, after presenting the frameworks of Geman et al., James and
T\&G in detail, we reformulate James'\textit{ SE} and \textit{VE}
in terms of the T\&G parameters.  
We also provide two  case studies  to highlight the
 importance of the theoretical derivations established in this paper.

The links established in this study not only provide insights to understanding classification
performance, but also lead to closed form expressions of classification B\&V  for determining when and
how performance improves or deteriorates. 
The findings are expected to be useful to the pattern recognition and machine learning community, as the
B\&V decomposition of $0/1$ loss function and the B\&V trade-off are tools employed by many researchers. A few examples of recent studies addressing the issues raised in this paper can be given as follows: In \cite{fumera08}, a theoretical and experimental analysis of the T\&G framework is utilised for analysing the performance of linear combination rules and bagging ensembles. In \cite{merentitis14}, the first attempt at a unified framework for hyperspectral image classification from a B\&V decomposition point of view is made, while considering all steps of the classification process: feature extraction, feature selection, classification, and post-processing. The proposed decision tree ensemble in \cite{zhang15}, is shown to achieve increased accuracy by guaranteeing a significant reduction in variance and a lower bias compared to the standard ensemble methods.  The B\&V decomposition used in \cite{younsi16} shows how their proposed ensemble classifiers decrease error due to a reduction in bias, a reduction in unbiased variance, an increase in biased variance or, a combination of these factors. In \cite{narassiguin16}, an extensive experimental comparison of the B\&V trade-off is made  for sixteen ensemble methods. A useful discussion about the B\&V decomposition problem is provided in \cite{mahajan17}, and a novel attempt is made at estimating B\&V. In \cite{nguyen18}, the proposed novel classifier combining rule is shown to be superior to other fixed combining rules by demonstrating that it has noticeably lower bias and slightly higher variance. In \cite{lerman18}, the B\&V decomposition is used in cryptographic systems to determine the impacts of the error rate of template and stochastic attacks, and also to extract the best profiled attack.  In order to achieve a trade-off between fairness and accuracy in sensitive applications such as healthcare or criminal justice, in \cite{chen18} cost-based metrics of discrimination are decomposed into bias, variance, and noise, and methods are proposed for estimating and reducing each term.

\section{Bias-Variance Frameworks\label{sec:Bias-Variance-Background}}

The bias and variance decomposition of the prediction error was initially
performed by  
\cite{DBLP:journals/neco/GemanBD92}  in the regression setting using squared-error loss,      and can be summarised as follows.

Let the estimate of the response, $Y$,
for each input pattern (feature vector) $X$ be $\hat{Y}$, which is a variable depending on the training
set. Different training sets consist of different patterns sampled from the same underlying distribution, or same patterns represented by different modalities. Given a particular pattern $X$, the effectiveness
of the prediction can be measured using the mean squared
error as 
\begin{alignat}{1}
E_{Y,\hat{Y}}\left[\left(Y-\hat{Y}\right){}^{2}|X\right] & =E_{\hat{Y}}\left[\left(\hat{Y}-E_{\hat{Y}}\left[\hat{Y}|X\right]\right)^{2}|X\right]
  +\left(E_{\hat{Y}}\left[\hat{Y}|X\right]-E_{Y}\left[Y|X\right]\right)^{2}+E_{Y}\left[\left(Y-E_{Y}[Y|X]\right)^{2}|X\right]\label{eq:gemanAsil}
\end{alignat}

In Eq. \ref{eq:gemanAsil}, the first term on the r.h.s., indicating
the expected difference between the estimate and its expectation,
is named as \textit{variance} of the predictor; whereas the second
term, indicating the squared difference between the regression and
the expected estimation as its \textit{bias}. The third term
is the variance of the response, which can be referred to as the \textit{irreducible
noise}.

Subsequently,  \cite{Breiman98arcingclassifiers}, 
  \cite{DBLP:conf/icml/KahaviW96},   
\cite{DBLP:conf/icml/KongD95},  \cite{DBLP:journals/datamine/Friedman97},
 \cite{DBLP:journals/neco/Wolpert97},  \cite{journals/neco/Heskes98},
 \cite{Tibshirani96biasvariance},  \cite{DBLP:conf/aaai/Domingos00}
and  \cite{DBLP:journals/ml/James03} have extended the
analysis for the classification setting; however, a universally accepted
framework still does not exist. Although all frameworks aim at preserving
the meanings of the standard B\&V terms and the additive property
of the error decomposition, they have difficulties maintaining other crucial 
B\&V characteristics and/or bring about undesirable properties, such that: 1) Dietterich and Kong allow
 the existence of negative variance, and moreover,
allow the Bayes classifier to have positive bias.  2) For each input
pattern, Breiman separates the predictors into two sets as biased
and unbiased; and accordingly, each test pattern has solely
bias or variance. 3) Kohavi
and Wolpert assign a non-zero bias to the Bayes classifier. 4) The definitions of
Tibshirani, Heskes and Breiman are difficult to generalise and extend
to general loss functions. 

Among all definitions, the framework of James overcomes
the listed disadvantages, and offers advantages in the sense 
of characterising consistent B\&V decompositions for arbitrary loss
functions. This framework is investigated in detail in Section \ref{subsec:Bias-and-VarianceJames}. Note that in \cite{DBLP:journals/datamine/Friedman97,DBLP:journals/ml/James03,Kuncheva:2004:CPC:975251}, the links between the majority of these definitions have been established under unified representations.
%
%
%

\subsection{Bias-Variance Analysis of James \label{subsec:Bias-and-VarianceJames}}

\cite{DBLP:journals/ml/James03} extends the B\&V analysis, originally
proposed for squared error under regression setting by 
\cite{DBLP:journals/neco/GemanBD92}, to all symmetric loss functions including $0/1$. 
 James  argues that  it is not possible
to both preserve the additive property of the prediction error decomposition
and maintain the terms' standard meanings. Hence,
in his decomposition, two sets of formulations are proposed to satisfy
these requirements separately. First, by adhering to the original
meanings, the bias is defined as the average distance
between the systematic parts of the response and the predictor; and
the variance as the variability of the predictor. Second, in addition
to the bias and variance terms, the notions of \textit{systematic
effect (SE)} and \textit{variance effect (VE)} are proposed. These
new terms satisfy the additive error decomposition for all symmetric
loss functions,
and characterise the effects of bias and variance on the
prediction error. For example, having (positive) variance might actually
trigger a reduction in the prediction error and hence bring about
a negative \textit{VE}. It should be noted that in the standard setting,
under squared error loss, \textit{SE} and \textit{VE} simplify into
bias and variance respectively, satisfying the additive decomposition.

Using the B\&V of the standard setting from Eq. \ref{eq:gemanAsil}, the generalised B\&V
of an estimator $\hat{Y}$ for a given pattern $X$ is given by

\begin{align}
Bias\left(\hat{Y}_{X}\right) =L(SY_{X},\,S\hat{Y}_{X})  \quad \textrm{and} \quad
Var\left(\hat{Y}_{X}\right) =E_{\hat{Y}}[L(\hat{Y}_{X},\,S\hat{Y}_{X})]\label{eq:biasvarJames}
\end{align}
where $L(u,w)$ is the loss when $w$ is used in predicting $u$, and $S\hat{Y}$ and $SY$ are called the systematic parts of $\hat{Y}$
and $Y$, respectively (Note that we will drop the conditioning on
the input pattern $X$ for simplicity). For squared error loss ($L_{S}$),
we know that $S\hat{Y}=E_{\hat{Y}}[\hat{Y}]$, which is the
regression or the best predictor of $\hat{Y}$ under $L_{S}$
\cite{DBLP:journals/neco/GemanBD92}. Hence, during the extension
for general loss functions ($L$) as given in Eq. \ref{eq:biasvarJames},
$S\hat{Y}$ can be re-written as $S\hat{Y}=\arg\min_{w}E_{\hat{Y}}[L(\hat{Y},w)]$,
and likewise, $SY=\arg\min_{w}E_{Y}[L(Y,w)]$.

%
%

 The decomposition of the prediction error is interpreted
  as the sum of three
terms: the variance of the response, the systematic (bias) effect
(\textit{SE}) and the variance effect (\textit{VE}): 
\begin{gather}
E_{Y,\hat{Y}}\left[L\left(Y,\hat{Y}\right)\right]=Var(Y)+SE(\hat{Y},Y)+VE(\hat{Y},Y)\label{eq:jamesDecomp}
\end{gather}
where $SE(\hat{Y},Y)=E_{Y}[L(Y,S\hat{Y})-L(Y,SY)]$, and $VE(\hat{Y},Y)=E_{Y,\hat{Y}}[L(Y,\hat{Y})-L(Y,S\hat{Y})]$.


For the specific case of classification with $0/1$ loss,
we have $L(u,w)=I(u\neq w)$, where $I(q)$ is the indicator function. Assume
that we are working on a $k$ class problem, such that instead of
operating in a continuous domain as in regression, $Y$ takes on discrete
values: $Y\:\epsilon\:\{\omega_{1},\omega_{2},\omega_{3},...,\omega_{k}\}$.
If we define 
\begin{align}
P(\omega_{i}|X) & =p(Y=\omega_{i}|X)  \quad \textrm{and} \quad
\hat{P}(\omega_{i}|X)  =p(\hat{Y}=\omega_{i}|X)\label{eq:of14}
\end{align}
where $P$ is the posterior probability distribution function for
the response, $Y$, and $\hat{P}$ is the distribution function obtained
over multiple training sets for the estimation, $\hat{Y}$. Then,
\begin{align}
SY & =\arg\min_{\omega_{i}}E_{Y}[I(Y\neq\omega_{i})|X]=\arg\max_{\omega_{i}}P(\omega_{i}|X)\nonumber \\
S\hat{Y} & =\arg\max_{\omega_{i}}\hat{P}(\omega_{i}|X)_{.}\label{eq:of13}
\end{align}
It can be observed in Eq. \ref{eq:of13} that $SY$ is the class assigned by the Bayes rule; and $S\hat{Y}$
can be interpreted as the aggregate classifier decision. Therefore

\begin{align}
Bias_{J}(\hat{Y}) & =I(S\hat{Y}\neq SY|X)\nonumber \\
Var_{J}(Y) & =p(Y\neq SY|X)=1-\max P(\omega_{i}|X)\nonumber \\
Var_{J}(\hat{Y}) & =p(\hat{Y}\neq S\hat{Y}|X)=1-\max\hat{P}(\omega_{i}|X)\nonumber \\
SE(\hat{Y},Y) & =p(Y\neq S\hat{Y}|X)-p(Y\neq SY|X)=P(SY|X)-P(S\hat{Y}|X)\nonumber \\
VE(\hat{Y},Y) & =p(Y\neq\hat{Y}|X)-p(Y\neq S\hat{Y}|X)
 =P(S\hat{Y}|X)-\sum_{i}P(\omega_{i}|X)\hat{P}(\omega_{i}|X).\label{eq:JamesToplu}
\end{align}
Note that, the irreducible error, $Var_{J}(Y)$, is equal to the Bayes
error.

\subsection{Tumer and Ghosh Model (T\&G Model) \label{subsec:Tumer-and-GhoshMain} }

In contrast to the other B\&V classification frameworks, Tumer and Ghosh
(T\&G) provide an analytic model \cite{DBLP:journals/pr/TumerGOriginal,TumerGhoshCorrelated,tumerGhoshSon}
by utilising the class posterior probability distributions of
the data and their estimates given by the predictors, assuming these
distributions are known. By decomposing the squared estimation errors belonging to these distributions using Geman's formulation, the added classification error (on top of the Bayes error) for a non-optimal learning algorithm is measured. It is important to note that this framework does not propose new terminologies of bias and variance to analyse the classification error under $0/1$ loss, rather uses the standard terminology in an unconventional way to address the new setting.
 The analysis is confined to the data points (patterns)
$X\in R$ instead of $X\in R^{n}$, but the conclusions derived are
expected to hold for $X\in R^{n}$.

\begin{figure}
\centering
\includegraphics[scale=0.60]{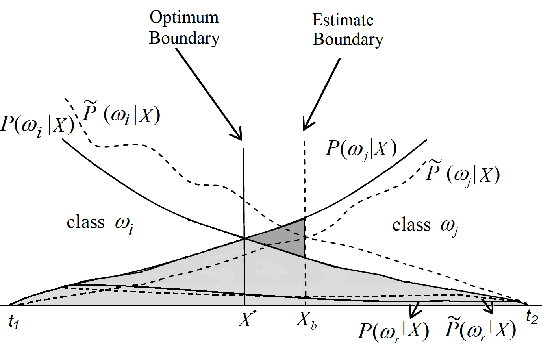}
\caption{Plot of real and estimated posterior probability distributions
with  boundaries}
\label{tgFigure}  
\end{figure}





The T\&G theory assumes that in a localised decision (transition)
region, only two classes possess significant posterior probabilities,
which will leave the contributions of other classes as negligible.
Consider a multi-class problem and the posterior probability functions
belonging to classes $\omega_{i}$ and $\omega_{j}$ for a data point
$X$, given as $P(\omega_{i}|X)$ and $P(\omega_{j}|X)$ as illustrated
in Figure \ref{tgFigure}. The summation of the posterior probabilities
for the rest of the classes can be referred to as $P(\omega_{r}|X)$.
An instance of the imperfect predictor, namely a base classifier which
is trained on one realisation of a given data set, would approximate
each posterior probability with an error (depicted by dashed lines
in Figure \ref{tgFigure}) $\tilde{P}(\omega_{i}|X)=P(\omega_{i}|X)+\epsilon_{i}(X)$
where $\epsilon_{i}(X)$ is a random variable denoting the 
approximation (prediction) error for the class $\omega_{i}$ given
input $X$. 

The optimal decision boundary given by the Bayes classifier can be
depicted as the intersection line crossing through $X^{*}$ in Figure \ref{tgFigure}. It
is then possible to denote any pattern in the decision region in terms
of its distance to the optimal boundary, such that $X_{a}=X^{*}+a$,
and name the set of all possible distance values, $a$, in the decision
region as $A$. Note that the infimum and the supremum of the set 
$A$ are denoted as $t_{1}$ and $t_{2}$ in Figure \ref{tgFigure}. Hence, a pattern located at the estimate decision
boundary of a base classifier can be given by $X_{b}=X^{*}+b$. Therefore,
\begin{gather}
\tilde{P}(\omega_{i}|X^{*}+b)=\tilde{P}(\omega_{j}|X^{*}+b)\nonumber \\
P(\omega_{i}|X^{*}+b)+\epsilon_{i}(X_{b})=P(\omega_{j}|X^{*}+b)+\epsilon_{j}(X_{b}).\label{eq:of11}
\end{gather}
By making the assumption that the posterior probabilities are locally
monotonic functions around the decision boundaries in the transition
regions, a linear approximation can be given:
\begin{gather}
P(\omega_{i}|X^{*})+bP'(\omega_{i}|X^{*})+\epsilon_{i}(X_{b})=P(\omega_{j}|X^{*})+bP'(\omega_{j}|X^{*})+\epsilon_{j}(X_{b})\label{eq:appr}
\end{gather}
\medmuskip=3mu \thinmuskip=4mu \thickmuskip=5mu where $P'$ denotes
the derivative of $P$. Using $P(\omega_{i}|X^{*})=P(\omega_{j}|X^{*})$
to rewrite Eq. \ref{eq:appr},

\begin{gather}
b=\frac{\epsilon_{i}(X_{b})-\epsilon_{j}(X_{b})}{s}\label{eq:b}
\end{gather}
where $s=P'(\omega_{j}|X^{*})-P'(\omega_{i}|X^{*})$. Note again that $\epsilon_{i}(X_{b}), \epsilon_{j}(X_{b})$ and $b$ are random variables dependent
on the posterior probability estimations obtained from the given
learning algorithm. It is assumed that $\epsilon_{i}(X)$ can be decomposed
into a class-specific bias, $\beta_{i}$; and a zero-mean, $\left(\sigma_{i}\right)^{2}$
class-specific variance noise term, $n_{i}$, which has the same distribution
characteristics across all $X$: $\epsilon_{i}(X)=\beta_{i}+n_{i}$.
Thus,
\begin{gather}
b=\frac{n_{i}-n_{j}}{s}+\frac{\beta_{i}-\beta_{j}}{s}.\label{eq:b2}
\end{gather}

In Figure \ref{tgFigure}, the Bayes error region is depicted by the
areas shaded by light grey (note the double addition of the areas
under $P(\omega_{r}|X)$).
 The added error region (on top of the Bayes
error) associated with the given realisation of the non-optimal classifier
is the dark shaded area, which can be approximated
by a triangle whose height is $b$, and base length is given by
\begin{alignat}{1}
P(\omega_{j}|X^{*}+b)-P(\omega_{i}|X^{*}+b)  =P(\omega_{j}|X^{*})+b\tilde{P}(\omega_{j}|X^{*})-P(\omega_{i}|X^{*})-b\tilde{P}(\omega_{i}|X^{*})
  =bs.\label{eq:of10}
\end{alignat}
Accordingly, the area of this triangular region is equal to $b^{2}s/2$;
with  expected value, $R_{add}$:
\begin{align}
R_{add} & =\frac{s}{2}\intop_{b\epsilon A}b^{2}p(b)db\label{eq:Eadd1}
\end{align}
where $p(b)$ is the probability distribution function of $b$, which
is defined within the decision region $A$. It will now be shown that
$R_{add}$ can be written in terms of the B\&V associated
with the classifier estimation errors. For that, let us first define
the variance for the random variable $b$:


\begin{align}
\sigma_{b}^{2} & =\intop_{b\epsilon A}(b-E[b])^{2}p(b)db\
  =\intop_{b\epsilon A}b^{2}p(b)db-2E[b]\intop_{b\epsilon A}bp(b)db+(E[b])^{2}
  =\intop_{b\epsilon A}b^{2}p(b)db-(E[b])^{2}\label{eq:varTG1}
\end{align}

Eq. \ref{eq:varTG1} can be combined with Eq. \ref{eq:Eadd1} such
that 
\begin{align}
\frac{s}{2}\sigma_{b}^{2} & =\frac{s}{2}\left(\intop_{b\epsilon A}b^{2}p(b)db-(E[b])^{2}\right)=R_{add}-\frac{s}{2}E[b].\label{eq:of9}
\end{align}
Therefore,
\begin{align}
R_{add} & =\frac{s}{2}\left(\sigma_{b}^{2}+(E[b])^{2}\right)=\frac{s}{2}\left(\sigma_{b}^{2}+\beta^{2}\right)\label{eq:EaddImportant}
\end{align}
where $\sigma_{b}^{2}$ is the T\&G overall variance ($Var_{TG}=\sigma_{b}^{2})$
and $\beta$ is the T\&G overall bias ($Bias_{TG}=\beta$).

In Eq. \ref{eq:EaddImportant}, the expected added area is shown to
be a function of the standard B\&V in terms of the variable $b$,
which is itself a function of the classifier approximation errors
as given in Eq. \ref{eq:b}. Therefore, the overall B\&V utilised
in the T\&G model relates to the difference of the classifier approximation
errors. Using Eq. \ref{eq:b2}, the terms in Eq. \ref{eq:EaddImportant}
can now be further expanded. Initially, we have 
\begin{equation}
Bias_{TG}=\beta=E\left[b\right]=\frac{\beta_{i}-\beta_{j}}{s}\label{eq:Ebsingle}
\end{equation}
An alternative expression for $\sigma_{b}^{2}$ can be derived using Eq. \ref{eq:b2}:
\begin{align}
Var_{TG}=\sigma_{b}^{2} & =var\left(\frac{n_{i}-n_{j}}{s}\right)\label{eq:varbSinglePre}\\
 & =\frac{1}{s^{2}}E\left[(n_{i}-n_{j}-E(n_{i}-n_{j}))^{2}\right]\nonumber \\
 & =\frac{\left(\sigma_{i}\right)^{2}+\left(\sigma_{j}\right)^{2}-2cov\left(n_{i},n_{j}\right)}{s^{2}}\label{eq:varbSingle}
\end{align}
where $\left(\sigma_{i}\right)^{2}$ is the class-specific variance
of $n_{i}$. Finally, substituting Eq. \ref{eq:Ebsingle}
and Eq. \ref{eq:varbSingle} in Eq. \ref{eq:EaddImportant} leaves
us with 
\begin{align}
R_{add} & =\frac{\left(\sigma_{i}\right)^{2}+\left(\sigma_{j}\right)^{2}-2cov\left(n_{i},n_{j}\right)}{2s}+\frac{s\beta^{2}}{2}\label{eq:eaddSingle2}
\end{align}
which explicitly defines the expected added error region in terms
of the B\&V of the individual class approximation errors and is the
main result of the T\&G analysis.

As a summary, T\&G obtains an additive B\&V decomposition of the classification error because its underlying
assumption on the posterior estimates allows the added error to be expressed as a function of the location of the decision boundary, which is dependent on the estimation noise. This assumption, which allows the added error region to be expressed as a triangle, implicitly converts the classification error problem into a regression problem.
Refer to \cite{DBLP:journals/pr/TumerGOriginal,TumerGhoshCorrelated,tumerGhoshSon} for further details about the T\&G model, to \cite{Kuncheva:2004:CPC:975251} for an analysis of the shortcomings of the T\&G model assumptions,
and to \cite{biggio07,biggio09} for an extension of the T\&G model where assumptions such as the existence of the ideal boundary between the same classes in the vicinity of the estimate boundary are relaxed. 

\section{Connections Between the Frameworks of Geman et al., James and T\&G \label{sec:jamestg}}



In this study, we establish the
relation between the frameworks of Geman et al., James and T\&G  
in order to 
reveal when and how the classification B\&V affect
the prediction error, i.e. the components that trigger an increase
or  decrease in \textit{SE} and \textit{VE} are given in closed form. 
 Specifically, after setting up the connection between the T\&G model and the decomposition of Geman et al. in Section  \ref{subsec:ConnectionTgGeman}, we formulate
\textit{SE} and \textit{VE} of James' framework in terms of B\&V of
T\&G in Section \ref{subsec:ConnectionJamesTg}.
A discussion on the established connections between the two frameworks are given in Section \ref{subsec:Discussioniki}, followed by two case studies underlining the findings of this theoretical study in Section \ref{sec:case}.

\subsection{T\&G Model in terms of Bias-Variance of Geman et al. \label{subsec:ConnectionTgGeman}}

By decomposing the squared posterior probability prediction error,
$\epsilon_{i}(X)$, into terms of bias and a zero-mean noise for a
given pattern $X$ and class $\omega_{i}$, T\&G model is similar to
the formulations of Geman et al. given in the regression framework. 

From T\&G model we know that $\epsilon_{i}(X)=\beta_{i}+n_{i}$. Therefore,
\begin{align}
\beta_{i}^{2} & =\left(E\left[\epsilon_{i}(X)\right]\right)^{2}=\left(E\left[\tilde{P}(\omega_{i}|X)-P(\omega_{i}|X)\right]\right)^{2}.\label{eq:TGGeman1}
\end{align}
where $\beta_{i}$ is the class-specific bias for the class $\omega_{i}$.
For the class-specific variance:

\begin{align}
\sigma_{i}^{2} & =E\left[\left(n_{i}-E\left[n_{i}\right]\right)^{2}\right]\nonumber \\
& =E\left[\left(\left(\epsilon_{i}(X)-\beta_{i}\right)-E\left[\epsilon_{i}(X)-\beta_{i}\right]\right)^{2}\right]\nonumber \\
& =E\left[\left(\epsilon_{i}(X)-E\left[\epsilon_{i}(X)\right]\right)^{2}\right]\nonumber \\
& =E\left[\left(\left[P(\omega_{i}|X)-\tilde{P}(\omega_{i}|X)\right]-E\left[P(\omega_{i}|X)-\tilde{P}(\omega_{i}|X)\right]\right)^{2}\right]\nonumber \\
& =E\left[\left(\tilde{P}(\omega_{i}|X)-E\left[\tilde{P}(\omega_{i}|X)\right]\right)^{2}\right].\label{eq:TGGeman2}
\end{align}

In contrast to the original framework of Geman et al. and its extension for
classification by James, where the B\&V terms are defined for the
response and its estimator ($Y$ and $\tilde{Y}$), T\&G define the
B\&V on the estimate posterior probabilities produced by the prediction
functions for each class (using $P$ and $\tilde{P}$). Let us now
compare Eq. \ref{eq:TGGeman1} and Eq. \ref{eq:TGGeman2} from T\&G
formulation with the original B\&V definitions of Geman et al. From Eq. \ref{eq:gemanAsil}, we have, 

\begin{align}
Bias(\hat{Y}) & =\left(E_{\hat{Y}}\left[\hat{Y}|X\right]-E_{Y}\left[Y|X\right]\right)^{2}\nonumber \\
Var(\hat{Y}) & =E_{\tilde{Y}}\left[\left(\hat{Y}-E_{\hat{Y}}\left[\hat{Y}|X\right]\right)^{2}|X\right].\label{eq:gemanbiasvar}
\end{align}
By interchanging $Y|X$ given in Eq. \ref{eq:gemanbiasvar} with $P(\omega_{i}|X)$,
and \textbf{$\tilde{P}(\omega_{i}|X)$ }with $\hat{Y}|X$, it can
be observed that the class-specific B\&V terms of T\&G as formulated
in Eq. \ref{eq:TGGeman1} and Eq. \ref{eq:TGGeman2} can be obtained.
Remember that the relation of the class-specific terms to the T\&G
overall B\&V are provided in Eq. \ref{eq:Ebsingle} and Eq. \ref{eq:varbSingle}.
Note that 
although  $Y|X$ given in the Geman framework is a random variable, its T\&G substitute $P(\omega_{i}|X)$ is 
a constant. Also, the class-specific bias of T\&G is not squared ($\beta_{i}$
instead of $\beta_{i}^{2}$); leading to the possibility of a positive
or negative value.

\subsection{Bias-Variance Analysis of James using T\&G Model\label{subsec:ConnectionJamesTg}}

While establishing the connection between the B\&V decompositions
of James and T\&G, we will still be relying on the assumption made
in T\&G model that the added error region for a multiclass problem
would mainly be composed of the contributions of two classes, as specified in Section \ref{subsec:Tumer-and-GhoshMain}. Remembering that $a$ is the distance between the pattern $X_{a}$
located at $X^{*}+a$ and $X^{*}$, Eq. \ref{eq:jamesDecomp} can
explicitly be written for all patterns lying within the decision region
$A$ (i.e.  $\forall a, t_{1}<a<t_{2}$), such that  

\begin{align}
\intop_{a\epsilon A}E_{Y,\hat{Y}}\left[L\left(Y_{X_{a}},\hat{Y}_{X_{a}}\right)\right]\,da & =\intop_{a\epsilon A}Var_{J}(Y_{X_{a}})da\nonumber \\
& +\intop_{a\epsilon A}SE(\hat{Y}_{X_{a}},Y_{X_{a}})da+\intop_{a\epsilon A}VE(\hat{Y}_{X_{a}},Y_{X_{a}})da.\label{eq:jamesDecWithA}
\end{align}

Since under $0/1$ loss $Var_{J}(Y_{X_{a}})$ is equal to the Bayes error
for the pattern $X_{a}$, Eq. \ref{eq:jamesDecWithA} can be re-written
as

\begin{align}
R_{add} & =\intop_{a\epsilon A}SE(\hat{Y}_{X_{a}},Y_{X_{a}})da+\intop_{a\epsilon A}VE(\hat{Y}_{X_{a}},Y_{X_{a}})da.\label{eq:Eadd}
\end{align}
For patterns outside the decision region, \textit{SE} and \textit{VE}
are zero, making the total expected error 
equal to the Bayes error, and leaving $R_{add}$ as zero.

In Section \ref{subsec:Calculation-of-SE} and Section \ref{subsec:Calculation-of-VE},
we will derive \textit{SE} and \textit{VE} using
T\&G terminology. However, first, we dedicate
Section \ref{subsec:bilmemne} for the expansions of the probability
terms $P$ and $\hat{P}$ to be later used in the \textit{SE} and
\textit{VE} calculations.

\subsubsection{Probability Distribution Analysis \label{subsec:bilmemne}}

Using Figure~~\ref{tgFigure} and Eq.~\ref{eq:of10}, for any $a>0$, 
\begin{equation}
P(\omega_{j}|X_{a})-P(\omega_{i}|X_{a})=sa\label{eq:as}
\end{equation}
where $\omega_{j}$ is the Bayes class. If the
Bayes error rate for $X_{a}$ is expressed by $z(X_{a})$, then

\begin{equation}
P(\omega_{j}|X_{a})=1-z(X_{a}).\label{eq:bayes}
\end{equation}
After denoting the summation of the posterior probabilities belonging
to classes with negligible contributions (which are considered as
noise) by $P(\omega_{r}|X)=\eta(X_{a})$, we also have 
\begin{equation}
P(\omega_{i}|X_{a})=z(X_{a})-\eta(X_{a}).\label{eq:secondMaj}
\end{equation}
Eq. \ref{eq:secondMaj} can be used to rewrite Eq. \ref{eq:as} such
that 
\begin{equation}
P(\omega_{j}|X_{a})=sa+\left(z(X_{a})-\eta(X_{a})\right).\label{eq: asArti}
\end{equation}

Similarly, for $a<0$ we have

\begin{equation}
P(\omega_{i}|X_{a})-P(\omega_{j}|X_{a})=-sa\label{eq:eksias}
\end{equation}
for which the Bayes class is $\omega_{i}$. The remaining calculations
for $a<0$ follow in accordance with those given for $a>0$ above,
and a summary of the findings is provided in Table~\ref{vetable}-a.
Here, for each $a$ value, the Bayes class is indicated as
$SY$, and the second dominant class is named as $SZ$.  

In Table~\ref{vetable}-b, the distribution function obtained over
training sets for the prediction, $\hat{P}$, is analysed. We know
that the assignment of a pattern $X_{a}$ into the classes $\omega_{i}$
or $\omega_{j}$ depends on the location of the estimation boundary
$X_{b}=x^{*}+b$, with $b$ being a random variable. Hence, while
calculating $\hat{P}$, the distribution of $b$ is utilised. From Figure~\ref{tgFigure}, for the
input pattern $X_{a}$ to be assigned to the class $\omega_{i}$ by
a classifier $c_{1}$, the boundary of the classifier has to be located
at $b_{c1}>a$. Hence, the probability of the aggregate decision for
$X_{a}$ being $\omega_{i}$ is equal to the sum of the probabilities
of all such boundaries, i.e. $\hat{P}(\omega_{i}|X_{a})=\intop_{b=a}^{t_{2}}p(b)db$. Similar analysis
follows for $\hat{P}(\omega_{j}|X_{a})$. As for $\hat{P}(\omega_{r}|X_{a})=0$,
we know that due to the T\&G assumption of having only two predominant
classes within $A$, the classifiers are not expected to assign a
pattern into any other class.

\begin{table}
\centering%
\begin{tabular}{|l|c|c|}
\hline 
& $a<0$  & $a>0$\tabularnewline
\hline 
\hline 
\textbf{$P(\omega_{i}|X_{a})$}  & $\begin{array}{c}
1-z(X_{a})\\
=-sa+\left(z(X_{a})-\eta(X_{a})\right)
\end{array}$  & $z(X_{a})-\eta(X_{a})$\tabularnewline
\hline 
$P(\omega_{j}|X_{a})$  & $z(X_{a})-\eta(X_{a})$  & $\begin{array}{c}
1-z(X_{a})\\
=sa+\left(z(X_{a})-\eta(X_{a})\right)
\end{array}$\tabularnewline
\hline 
$P(\omega_{r}|X_{a})$  & $\eta(X_{a})$  & $\eta(X_{a})$\tabularnewline
\hline 
$SY$  & $\omega_{i}$  & $\omega_{j}$\tabularnewline
\hline 
$SZ$  & $\omega_{j}$  & $\omega_{i}$\tabularnewline
\hline 
\end{tabular}\\
\centering(a)\\
$\;\;$\\

\centering%
\begin{tabular}{|c|c|}
\hline 
& $\forall a$\tabularnewline
\hline 
\hline 
$\hat{P}(\omega_{i}|X_{a})$  & $\intop_{b=a}^{t_{2}}p(b)db$\tabularnewline
\hline 
$\hat{P}(\omega_{j}|X_{a})$  & $\intop_{b=t_{1}}^{a}p(b)db$\tabularnewline
\hline 
$\hat{P}(\omega_{r}|X_{a})$  & $0$\tabularnewline
\hline 
\end{tabular}

\centering(b)

\caption{Derivations of $P$ and $\hat{P}$ for different $a$ values.}

\label{vetable} 
\end{table}

Making use of the expansions provided in this section, \textit{SE}
and \textit{VE} will be derived employing the T\&G terminology in
the following sections. During the derivations, $da$ will be dropped
for simplicity.

\subsubsection{Calculation of \textit{SE} \label{subsec:Calculation-of-SE}}

Using Eq. \ref{eq:JamesToplu}, we have,

\begin{align}
\intop_{a\epsilon A}SE(\hat{Y}_{X_{a}},Y_{X_{a}}) & =\intop_{a\epsilon A}P(SY|X_{a})-\intop_{a\epsilon A}P(S\hat{Y}|X_{a}).\label{eq:SEzeroOne}
\end{align}
We will now analyse the terms within Eq. \ref{eq:SEzeroOne} separately.
Since $SY=\arg\max_{\omega_{i}}P(\omega_{i}|X_{a})$ under $0/1$ loss
(see Eq. \ref{eq:of13}), $P(SY|X_{a})$ denotes 
the posterior probability of the class that the Bayes classifier
assigns for $X_{a}$. Thus,

\begin{align}
\intop_{a\epsilon A}P(SY|X_{a}) & =\intop_{a=t_{1}}^{0}P(\omega_{i}|X_{a})+\intop_{a=0}^{t_{2}}P(\omega_{j}|X_{a}).\label{eq:PYSY}
\end{align}

The calculation of $P(S\hat{Y}|X_{a})$ on the other hand is not as
trivial. From Eq. \ref{eq:of13}, $S\hat{Y}=\arg\max_{\omega_{i}}\hat{Y}(\omega_{i}|X_{a})$,
which suggests that $P(S\hat{Y}|X_{a})$ stands for the underlying
posterior probability of the most probable class according to the
base classifier decisions. To analyse $S\hat{Y}$, let us first determine
when it is equal to the Bayes decision class, $SY$. It can be observed
from Table \ref{vetable}-a that for $a>0$, $SY=\omega_{j}$. Hence,

\begin{equation}
\begin{cases}
S\hat{Y}=SY & \text{ if }\hat{P}(\omega_{j}|X_{a})>\hat{P}(\omega_{i}|X_{a})\\
& \text{ }\,\,\implies\intop_{b=t_{1}}^{a}p(b)db\}>\intop_{b=a}^{t_{2}}p(b)db.
\end{cases}\text{ }\label{eq:PYSYt1}
\end{equation}
Conversely, for patterns with $a<0$,

\begin{align}
\begin{cases}
S\hat{Y}=SY & \text{ if }\hat{P}(\omega_{i}|X_{a})>\hat{P}(\omega_{j}|X_{a})\\
& \text{ }\,\,\implies\intop_{b=t_{1}}^{a}p(b)db<\intop_{b=a}^{t_{2}}p(b)db.
\end{cases}\label{eq:PYSYt2}
\end{align}

The point $m$ where $\intop_{b=t_{1}}^{m}p(b)db=\intop_{b=m}^{t_{2}}p(b)db=0.5$
is the median of the probability distribution of $b$. For the case
of $0<m$, Eq. \ref{eq:PYSYt1} and Eq. \ref{eq:PYSYt2} can be simplified
such that

\begin{align}
\begin{cases}
S\hat{Y}=SY & \text{ if\thinspace\ }0<m<a \thinspace\,\,\, or \,\,\,\,\thinspace a<0<m
\end{cases}\,\,\,\,\,\label{eq:PYSYt1-1}
\end{align}
Bearing in mind the assumption that within the localised decision
region only two classes are likely to have considerable posterior
probabilities, we have $S\hat{Y}=SZ$ (second dominant class) for
all other $a$ than those defined in Eq. \ref{eq:PYSYt1-1}. $P(S\hat{Y}|X_{a})$
can now be reformulated as

\begin{alignat}{1}
\intop_{a\epsilon A}P(S\hat{Y}|X_{a}) & =\intop_{a<0}P(SY|X_{a})\nonumber \\
& +\intop_{0<a<m}P(SZ|X_{a})+\intop_{a>m}P(SY|X_{a}).\label{eq:PYSYt3}
\end{alignat}

Using Table \ref{vetable}-a, Eq. \ref{eq:PYSYt3} can be expanded
to 
\begin{alignat}{1}
\intop_{a\epsilon A}P(S\hat{Y}|X_{a}) & =\intop_{a=t_{1}}^{0}P(\omega_{i}|X_{a})\nonumber \\
& +\intop_{a=0}^{m}P(\omega_{i}|X_{a})+\intop_{a=m}^{t_{2}}P(\omega_{j}|X_{a}).\label{eq:PYSYtSon}
\end{alignat}
Similar analysis shows that the result of Eq. \ref{eq:PYSYtSon} also holds
for $m<0$.

Combining the derivations for $P_{SY}$ and $P_{S\hat{Y}}$ from Eq.
\ref{eq:PYSY} and \ref{eq:PYSYtSon}, and referring to Table \ref{vetable}-a,
Eq. \ref{eq:SEzeroOne} becomes

\begin{align}
\intop_{a\epsilon A}SE(\hat{Y}_{X_{a}},Y_{X_{a}}) & =\intop_{a=0}^{t_{2}}P(\omega_{j}|X_{a})-\intop_{a=0}^{m}P(\omega_{i}|X_{a})-\intop_{a=m}^{t_{2}}P(\omega_{j}|X_{a})\nonumber \\
& =\intop_{a=0}^{m}\left(P(\omega_{j}|X_{a})-P(\omega_{i}|X_{a})\right)\nonumber \\
& =\intop_{a=0}^{m}sa=\frac{m^{2}s}{2}.\label{eq:SESon}
\end{align}
Note that in Eq. \ref{eq:SESon}, $s$ is positive.

\subsubsection{Calculation of \textit{VE \label{subsec:Calculation-of-VE}}}

Using the given definition of \textit{VE} from Eq. \ref{eq:JamesToplu},
we have

\begin{align}
\intop_{a\epsilon A}VE(\hat{Y}_{X_{a}},Y_{X_{a}}) & =\intop_{a\epsilon A}P(S\hat{Y}|X_{a})-\intop_{a\epsilon A}\left[\sum_{i}P(\omega_{i}|X_{a})\hat{P}(\omega_{i}|X_{a})\right].\label{eq:VE}
\end{align}

The first term of Eq. \ref{eq:VE}, $\intop_{a\epsilon A}P(S\hat{Y}|X_{a})$,
has been derived during the calculation of \textit{SE} in Eq. \ref{eq:PYSYtSon}.
Using Table \ref{vetable}-a, Eq. \ref{eq:PYSYtSon} can further be
expanded to take the following form: 
\begin{alignat}{1}
\intop_{a\epsilon A}P(S\hat{Y}|X_{a}) & =\intop_{a=t_{1}}^{0}\left(1-z(X_{a})\right)+\intop_{a=0}^{m}\left(z(X_{a})-\eta(X_{a})\right)+\intop_{a=m}^{t_{2}}\left(1-z(X_{a})\right)\nonumber \\
& =\intop_{a=t_{1}}^{0}\left[-sa+\left(z(X_{a})-\eta(X_{a})\right)\right]+\intop_{a=0}^{m}\left(z(X_{a})-\eta(X_{a})\right)\nonumber \\
& +\intop_{a=m}^{t_{2}}\left[sa+\left(z(X_{a})-\eta(X_{a})\right)\right]\nonumber \\
& =\intop_{a\epsilon A}\left(z(X_{a})-\eta(X_{a})\right)+\frac{s}{2}\left[(t_{1}^{2}+t_{2}^{2})-m^{2}\right]\label{eq:PYSYtExpand}
\end{alignat}
For the second term of Eq. \ref{eq:VE}, Table \ref{vetable} is
used again (For convenience,
$da$ and $db$ will be dropped in the intermediate steps.): 
\begin{alignat}{1}
\intop_{a\epsilon A}\left[\sum_{i}P(\omega_{i}|X_{a})\hat{P}(\omega_{i}|X_{a})\right]da & =\intop_{a=t_{1}}^{0}\left[-sa+\left(z(X_{a})-\eta(X_{a})\right)\right]\intop_{b=a}^{t_{2}}p(b)dbda+\intop_{a=t_{1}}^{0}\left(z(X_{a})-\eta(X_{a})\right)\intop_{b=t_{1}}^{a}p(b)dbda\nonumber \\
& +\intop_{a=0}^{t_{2}}\left(z(X_{a})-\eta(X_{a})\right)\intop_{b=a}^{t_{2}}p(b)dbda+\intop_{a=0}^{t_{2}}\left[sa+\left(z(X_{a})-\eta(X_{a})\right)\right]\intop_{b=t_{1}}^{a}p(b)dbda\nonumber \\
& =\intop_{a=t_{1}}^{t_{2}}\left(z(X_{a})-\eta(X_{a})\right)+\intop_{a=t_{1}}^{0}-sa\intop_{b=a}^{t_{2}}p(b)+\intop_{a=0}^{t_{2}}sa\intop_{b=t_{1}}^{a}p(b)\nonumber \\
& =\intop_{a=t_{1}}^{t_{2}}\left(z(X_{a})-\eta(X_{a})\right)+s\text{\ensuremath{\Bigg[}}\intop_{b=0}^{t_{2}}\intop_{a=b}^{t_{2}}ap(b)+\intop_{b=t_{1}}^{0}\intop_{a=0}^{t_{2}}ap(b)+\intop_{b=0}^{t_{2}}\intop_{a=t_{1}}^{0}-ap(b)+\intop_{b=t_{1}}^{0}\intop_{a=t_{1}}^{b}-ap(b)\text{\text{\text{\ensuremath{\Bigg]}}}}\nonumber \\
& =\intop_{a\epsilon A}\left(z(X_{a})-\eta(X_{a})\right)+\intop_{b\epsilon A}\frac{s}{2}\left[(t_{1}^{2}+t_{2}^{2})-b^{2}\right]p(b)db.\label{eq:uzunC}
\end{alignat}
Using Eq. \ref{eq:Eadd1} for added error from Section \ref{subsec:Tumer-and-GhoshMain},
Eq. \ref{eq:uzunC} can further be simplified.

\begin{alignat}{1}
\intop_{a\epsilon A}\left[\sum_{i}P(\omega_{i}|X_{a})\hat{P}(\omega_{i}|X_{a})\right] & =\intop_{a\epsilon A}\left(z(X_{a})-\eta(X_{a})\right)\nonumber \\
& +\frac{s(t_{1}^{2}+t_{2}^{2})}{2}-R_{add}\label{eq:kisaC}
\end{alignat}
By combining the two terms derived in Eq. \ref{eq:PYSYtSon} and Eq.
\ref{eq:kisaC}, the formulation of \textit{VE} given in Eq. \ref{eq:VE} can finally be expanded as follows:

\begin{alignat}{1}
\intop_{a\epsilon A}VE(\hat{Y}_{X_{a}},Y_{X_{a}}) & =\intop_{a\epsilon A}\left(z(X_{a})-\eta(X_{a})\right)+\frac{s}{2}\left[(t_{1}^{2}+t_{2}^{2})-m^{2}\right]\nonumber \\
& -\intop_{a\epsilon A}\left(z(X_{a})-\eta(X_{a})\right)-\frac{s(t_{1}^{2}+t_{2}^{2})}{2}+R_{add}\nonumber \\
& =R_{add}-\frac{m^{2}s}{2}.\label{eq:VESon}
\end{alignat}
By writing $R_{add}$ in terms of the T\&G overall bias and variance
($\beta$ and $\sigma_{b}^{2}$) as given in Eq. \ref{eq:EaddImportant},
Eq. \ref{eq:VESon} can be reformulated as

\begin{align}
\intop_{a\epsilon A}VE(\hat{Y}_{X_{a}},Y_{X_{a}}) & =R_{add}-\frac{m^{2}s}{2}\nonumber \\
& =\frac{(\beta^{2}+\sigma_{b}^{2}-m^{2})s}{2}.\label{eq:VESon2}
\end{align}
where $s$ is positive.

The verification for the expected added error calculation provided
in Eq. \ref{eq:Eadd} can be obtained by adding \textit{SE} and \textit{VE}
given in Eq. \ref{eq:SESon} and \ref{eq:VESon} to give
\begin{align}
R_{add} & =\intop_{a\epsilon A}SE(\hat{Y}_{X_{a}},Y_{X_{a}})da+\intop_{a\epsilon A}VE(\hat{Y}_{X_{a}},Y_{X_{a}})da.\nonumber \\
& =\frac{m^{2}s}{2}+R_{add}-\frac{m^{2}s}{2}\nonumber \\
& =R_{add}.\label{eq:RaddChecksum}
\end{align}

\subsection{Discussion \label{subsec:Discussioniki} }


 The systematic effect (\textit{SE})
of James has been reformulated in terms of T\&G model parameters in Eq. \ref{eq:SESon},
which shows that the effect of classification bias on
 prediction performance depends on the median of the decision boundary
distribution belonging to the base classifiers: \textit{SE} increases
with the squared value of the median of the decision boundary distribution. In a symmetric
distribution like the normal distribution, the median is equal to
the mean and\textit{ SE} is directly linked to the squared T\&G
overall bias, $\beta$.

The T\&G overall bias term, which is
equal to the mean of the decision boundary distribution, ($\beta=E\left[b\right]$),
was shown in Eq. \ref{eq:Ebsingle} to be the difference between the class-specific biases of the
individual estimation errors  belonging to
the two classes of interest ($\beta_{i}-\beta_{j}$) scaled by a positive constant.
Therefore, any circumstance causing the absolute value of this difference
to decrease brings about decreased \textit{SE}.
Examples are when both estimation error biases,  $\beta_{i}$ and $\beta_{j}$,
are positive or negative in equal amounts. On the other hand, having
 $\beta_{i}$ and  $\beta_{j};$ of opposite sign  would create larger confusion
regions, shift the expected value of the decision boundaries ($E[b]$)
away from the Bayes boundary, and cause an increase in \textit{SE.}

 It has been demonstrated in Eq. \ref{eq:VESon2} that
\textit{VE} is directly related to the T\&G overall variance, but also depends
 on the shape of the decision boundary distribution. The latter
is represented by $(\beta^{2}-m^{2})$, which shows that when the difference between
the mean and the median of the decision boundary distribution increases,
\textit{VE} increases as well. On the other hand, for cases with symmetric boundary distributions,
\textit{VE} only depends on the T\&G overall variance, as the difference
 between the mean and the median of the boundary distribution
vanishes.

In Eq. \ref{eq:varbSinglePre}, the T\&G overall variance is
defined as the variance of the decision boundary distribution, which
is equal to the variance of the difference of estimation noise measured
on the two classes of interest scaled by a
positive constant. In Eq. \ref{eq:varbSingle}, this was shown to be
equal to the summation of the class-specific variance terms belonging
to the individual estimation errors ($(\sigma_{i})^{2}$ and $(\sigma_{j})^{2}$),
which are accompanied by a covariance term that vanishes for independent
classifier outputs. Hence, an increase in any of the
class-specific variance terms belonging to individual estimation errors
always causes an increase in the T\&G overall variance, and therefore
\textit{VE}.

\section{Case Studies \label{sec:case}}

In order to demonstrate the applicability of the T\&G and James frameworks on real problems and provide practical insight, we experimentally evaluate two example scenarios in Section~\ref{subsec:case1}-\ref{subsec:case2}, by using a benchmark data set. In Section~\ref{subsec:case1}, we use the T\&G theory to design more accurate ensembles and demonstrate the links established in this paper between the T\&G and James frameworks. In Section~\ref{subsec:case2}, we analyse the effect of James variance on classification performance by relating it to the T\&G model parameters. Since the T\&G model is based on assumptions such as the existence of only two dominant classes at the decision region and the single dimensionality of the data, we will assume that the theory is applicable to multiple dimensions, and make appropriate approximations during the calculations.

\subsection{Case Study 1 \label{subsec:case1}}

In this case study, we analyse and compare the performance of a single classifier and an ensemble by utilising the B\&V frameworks of  T\&G and James. The ensemble rule selected for this study is the average, which is the multiple classifier system with  mean combination. After summarising the theoretical derivations provided for the ensemble average under the T\&G model in \cite{DBLP:journals/pr/TumerGOriginal,TumerGhoshCorrelated,Kuncheva:2004:CPC:975251}, we provide an experimental analysis on a benchmark data set by employing both of the frameworks. 

First, we define $b_{ave}$ to be the decision boundary of the ensemble by using Eq. \ref{eq:b2}, such that 
\begin{equation}
b_{ave}=\frac{\bar{n_{i}}-\bar{n_{j}}}{s}+\frac{\bar{\beta_{i}}-\bar{\beta_{j}}}{s}\label{eq:of8}
\end{equation}
where $\bar{n_{i}}=\frac{1}{N}\sum_{m=1}^{N}n_{i}^{m}$ and $\bar{\beta_{i}}=\frac{1}{N}\sum_{m=1}^{N}\beta_{i}^{m}$, 
with $N$ being the total number of base classifiers. Here, $n_{i}^{m}$
and $\beta_{i}^{m}$ are noise and bias terms belonging to the 
base classifier $m$ of the ensemble for the class
$\omega_{i}$. From Eq.  \ref{eq:of8} 
\begin{eqnarray}
\beta^{ave}= & E[b_{ave}] & =\frac{\bar{\beta_{i}}-\bar{\beta_{j}}}{s} \label{eq:ebEns}
\end{eqnarray}
\begin{eqnarray}
(\sigma^{ave})^{2}  =  var\left(\frac{\bar{n_{i}}-\bar{n_{j}}}{s}\right)\
  =  \frac{1}{s^{2}} E[(\bar{n_{i}}-\bar{n_{j}})^{2}]
  =  \frac{1}{s^{2}}E[(\bar{n_{i}})^{2}+(\bar{n_{j}})^{2}-2\bar{n_{i}}\bar{n_{j}}].\label{eq:varens1}
\end{eqnarray}
Following the derivations of  \cite{Kuncheva:2004:CPC:975251} for the expansion of the terms in Eq.~\ref{eq:varens1}, the added error of the mean combination rule takes the following form:
\begin{eqnarray}
R_{add}^{ave} & = & \frac{s}{2}\left((\sigma^{ave})^{2}+(E[b_{ave}])^{2}\right)\nonumber 
 =\frac{1}{N^{2}}\left(\sum_{m=1}^{N}\frac{(\sigma_{i}^{m})^{2}+(\sigma_{j}^{m})^{2}-2cov(n_{i}^{m},n_{j}^{m})}{2s}\right)+\frac{s}{2}\left(\beta^{ave}\right)^{2}\nonumber \\
 &  & +\frac{1}{2N^{2}s}\left(\sum_{m=1}^{N}\sum_{n=1,n\neq m}^{N}cov(n_{i}^{m},n_{i}^{n})+cov(n_{j}^{m},n_{j}^{n})-2cov(n_{i}^{m},n_{j}^{n})\right).\label{eq:Eaddens}
\end{eqnarray}
T\&G analyse the added error for the specific case where there is no bias, under two assumptions:
\begin{enumerate}
\item The noise between classes are i.i.d. and have the same variance for
all $m$, i.e. $cov(n_{i}^{m},n_{j}^{m})=0$ and $\sigma_{i}^{m}=\sigma_{j}^{m}=\sigma$. This assumption leads 
to each classifier having the same expected added error, i.e. $R^{m}_{add}=R_{add}$, $\forall m$.
\item For different classes, noise between two classifiers are i.i.d, i.e.
$cov(n_{i}^{m},n_{j}^{n})=0$.
\end{enumerate}
It is shown in \cite{TumerGhoshCorrelated,Kuncheva:2004:CPC:975251} that under these assumptions, Eq. \ref{eq:Eaddens} can be
simplified into

\begin{equation}
R_{add}^{ave}=R_{add}\left(\frac{1+C(N-1)}{N}\right)\label{eq:Eaddens2}
\end{equation}
where $R_{add}$ is the added error of a single classifier, $C=\sum_{i=1}^{k}P_{i}C_{i}$, $P_{i}$ is the prior probability of the class $\omega_{i}$ and $C_{i}$ is the average correlation
coefficient among classifiers for this class. 

When identical classifiers are used for building the mean classifier ensemble,
i.e. $C=1$, Eq. \ref{eq:Eaddens2} takes the form $R_{add}^{ave}=R_{add}$. Thus, 
the difference between the average added errors
of the ensemble and the single classifier becomes zero, which means combining classifiers does not provide any benefit. On the other hand, when there is independence between the errors of any pair of base classifiers, i.e. 
$C=0$, the error of the ensemble becomes equal to the error of the single classifier divided by the number of classifiers.

\begin{figure}
\centering 
      \subfigure[ ]{\includegraphics[scale=0.4]{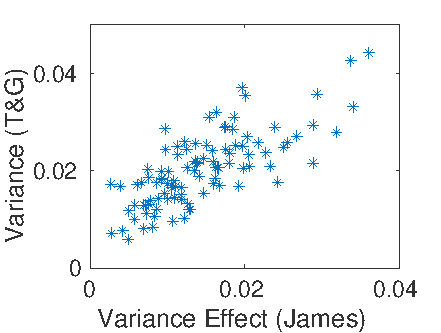}}
      \subfigure[ ]{\includegraphics[scale=0.4]{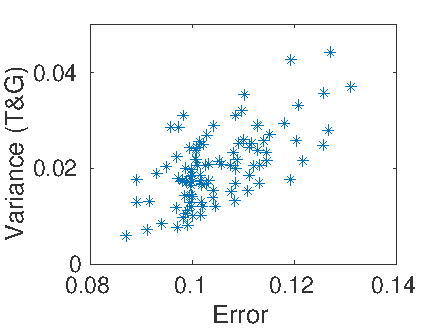}}
      \subfigure[ ]{\includegraphics[scale=0.4]{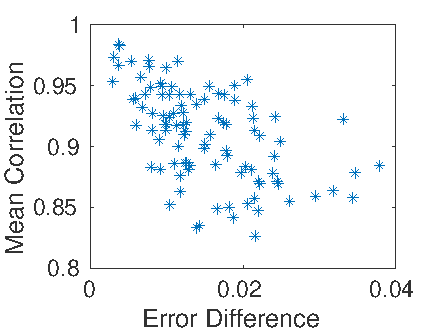}}
\caption{Example bias and variance analysis on single classifiers and ensembles, by using T\&G and James frameworks}
\label{case1fig}
\end{figure}

We aim to simulate the impact of variance in classification performance and the effect of correlation on the efficiency of the averager ensemble, by employing a benchmark data set, ``Image Segmentation'' \cite{Lichman13}. The data set consists of 7 classes and 19 dimensions, and comprises 210 training and 2100 test samples. To attain bias close to zero, we work with strong base classifiers, which are chosen as Neural Networks (NNs) with  a single hidden layer of 16 nodes and 8 epochs.

Firstly, 200 different groups of classifiers, each of which is formed of 6 NN classifiers with random initial weights, are created. For each group, the expected error rate, mean correlation coefficient ($C$), variance ($\sigma$) as given by T\&G,  $VE$ as formulated by James and     ensemble average  are recorded.  Note that T\&G make the assumption of fixed variance for all data and therefore the same expected added error for all classifiers. We try to comply with these assumptions by creating classifiers with similar variance when averaged over all test samples.

In Figure~\ref{case1fig} (a), in order to investigate the relation between the frameworks of T\&G and James, the T\&G variance for each of the 200 groups is plotted against its corresponding $VE$. A correlation coefficient of value 0.749 is measured between these variables, in line with the novel theoretical findings presented in Eq.~\ref{eq:VESon2}: After making the assumption that for strong base classifiers the mean ($\beta$) and the median ($m$) of the decision boundary approach zero, in Eq.~\ref{eq:VESon2} T\&G variance becomes directly related to James $VE$. Moreover, this assumption simplifies Eq.~\ref{eq:EaddImportant} such that the added error (and therefore the error) becomes a direct consequence of the T\&G variance. This can be visualised in Figure~\ref{case1fig} (b), where the correlation coefficient between the expected error rate of the base classifiers and the T\&G variance is 0.633.

Secondly, the behaviour of the correlation coefficient $C$ against the ensemble performance is depicted in Figure~\ref{case1fig} (c). The performance metric used here is the difference between the ensemble and the average base classifier accuracy. Supporting the T\&G findings given in Eq.~\ref{eq:Eaddens2}, where decrease in $C$ is shown to be related to gain in the average ensemble performance, the negative correlation coefficient between these two variables is measured as -0.512.

This case study validates the theory for powerful ensemble construction based on reduced average T\&G correlation between the base classifiers, on experimental data. Classification performance, T\&G and $VE$ have also been shown to be directly linked to each other under an approximately zero bias and zero median scenario. An important finding demonstrated here is that although T\&G framework depends on many assumptions, the theoretical outcomes generally hold in practice.

\subsection{Case Study 2 \label{subsec:case2}}

The aim of this case study is to visualise the effect of $VE$ on the prediction performance. It can be deduced from Eq.~\ref{eq:VESon2} that when the mean of the decision boundary distribution deviates from its median, $VE$ can actually take negative values. This circumstance, which is expected to occur when the classifier of interest is weak, implies that having James variance may have a positive impact on the generalisation error. In order to investigate, we use 50 NN classifiers and 5 node/epoch combinations on the Image Segmentation data set. The complexity of the NNs is decreased with the set of nodes and/or epochs: [(16/32), (8/8), (2/4), (1/1)]. As in Section~\ref{subsec:case1}, the classifier perturbation is achieved by using random initial weights.

\begin{figure}
\centering 
      {\includegraphics[scale=0.4]{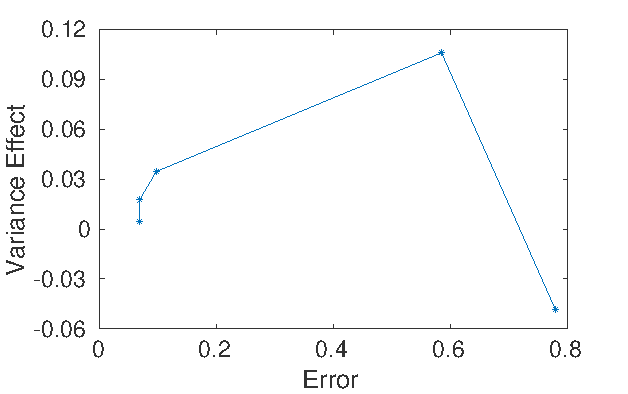}}
\caption{Variance Effect analysis for classifiers with decreasing complexity in direction shown}
\label{case2fig}
\end{figure}

In Figure~\ref{case2fig}, the error rates and $VE$ corresponding to the 5 combinations are plotted. As the error rate increases (i.e. as the complexity of the classifiers decreases), $VE$ can be observed to increase with the error until the error reaches a certain value ($0.6$). When the error rate is equal to $0.8$, the corresponding $VE$ drops below zero, which corresponds to a very weak classifier: a NN with 1 node and 1 epoch. The decision boundary for this classifier is expected to have a skewed distribution due to its inability to correctly identify the real boundary. Hence, having increased variance helps the classifier obtain the correct label sporadically by diverging from the aggregate false decision. In other words, we obtain a negative $VE$, which means that James variance actually boosts  the prediction performance. This explanation is also in line with the experimental evidence provided in \cite{DBLP:journals/ml/James03} and \cite{DBLP:series/sci/ZorWY11}. Finally, note that when the error is as low as $0.07$, that is for NNs with 16 nodes and 32 epochs, the $VE$ is expected to be directly related to T\&G variance similar to the case study in Section~\ref{subsec:case1}, as the mean and the medians of the decision boundary is expected to be similar and close to zero.

\section{Conclusions \label{sec:BiasVarConclusions} }

    
The relationships established in this paper not
only provide deeper insight into  classification theory, but also present
the effects of B\&V \textit{(SE} and \textit{VE}) defined by James
in a closed form, which is useful for explaining the
behaviours of these terms, such as when they cause performance increase or decrease.
 Understanding B\&V of T\&G in
terms of James terminology provides further advantages in
scenarios where it is not possible to measure the underlying class a
posteriori probability distributions.
Although the framework of James requires knowledge of the label
distribution for an input pattern, this information can usually be
approximated more accurately than the lower level probability distributions. This duality provides the user with the flexibility to be able to switch between the models while working on classifier design, and its use is highlighted on two case studies.

To date, the focus of research has been on classification B\&V and T\&G frameworks separately, but there has been no previous attempt to relate the two. It has to be remembered that James framework has been chosen in this paper as a representative among many classification B\&V formulations due to its advantages, and the established  T\&G linkage can be expanded to other B\&V frameworks by utilising the unified notations given in \cite{DBLP:journals/datamine/Friedman97,DBLP:journals/ml/James03,Kuncheva:2004:CPC:975251}.


There is no doubt that when there is a requirement to understand why a classifier or ensemble performs well, or when an experimental comparison of classifiers needs to be made, it is common practice for many researchers to refer to bias and variance analysis, 
examples of which are given in the Introduction.
Recently, \cite{belkin09} hypothesised that the reason deep networks do not over-fit is that, 
as complexity is increased, two bias-variance curves appear: variance begins to increase, and then decrease at the transition from the first to second bias/variance trade-off. As further work, we intend to investigate how the  two models described in this paper  may be used  to explain this over-fitting anomaly.




\bibliography{cemre,cemretw}
\bibliographystyle{ieeetr}

\end{document}